# A Rank-Based Similarity Metric for Word Embeddings

**Enrico Santus**[1], **Hongmin Wang**[2], **Emmanuele Chersoni**[3] and **Yue Zhang**[4]

esantus@mit.edu
hongmin_wang@cs.ucsb.edu
emmanuelechersoni@gmail.com
yue_zhang@sutd.edu.sg

[1] Computer Science and Artificial Intelligence Lab, MIT
[2] Department of Computer Science, University of California Santa Barbara
[3] Aix-Marseille University
[4] ISTD, Singapore University of Technology and Design

## Abstract

Word Embeddings (WE) have recently imposed themselves as a standard for representing word meaning in NLP. Semantic similarity between word pairs has become the most common evaluation benchmark for these representations, with vector cosine being typically used as the only similarity metric. In this paper, we report experiments with a *rank-based* metric for WE, which performs comparably to vector cosine in *similarity estimation* and outperforms it in the recently-introduced and challenging task of *outlier detection*, thus suggesting that rank-based measures can improve clustering quality.[1]

## 1 Introduction

> "All happy families resemble one another, but each unhappy family is unhappy in its own way."
> *Anna Karenina*, Leo Tolstoy

Distributional Semantic Models (DSMs) have received an increasing attention in the NLP community, as they constitute an efficient data-driven method for creating word representations and measuring their semantic similarity by computing their distance in the vector space (Turney and Pantel, 2010).

The most popular similarity metric in DSMs is the vector cosine. Compared to Euclidean distances, vector cosine scores are normalized on each dimension and hence are robust to the scaling effect. On the other hand, one limitation of this metric is that it regards each dimension equally, without taking into account the fact that some dimensions might be more relevant for characterizing the semantic content of a word. Such a limitation led to the introduction of alternative metrics based on feature ranking, which have been reported to outperform vector cosine in several similarity tasks (Santus et al., 2016a,b).

Recently, the focus of the research on word representations has been shifting onto the so-called *word embeddings* (WE), which are dense vectors obtained by means of neural network training that achieved significant improvements in several similarity-related tasks (Mikolov et al., 2013a; Baroni et al., 2014). Although the representation type of the embeddings was helpful for reducing the sparsity of traditional count vectors, their nature does not sensibly differ (Levy et al., 2015). Most research works involving WE still adopt vector cosine for similarity estimation, yet little experimentation has been done on alternative metrics for comparing dense representations (exceptions include Camacho-Collados et al. (2015)).

Some attempts to directly transfer rank-based measures from traditional DSMs to WE have faced difficulties (see, for example, Jebbara et al. (2018)). In this paper, we suggest a possible solution to this problem by adapting $APSyn$, a rank-based similarity metric originally proposed for sparse vectors (Santus et al., 2016b,a), to low-dimensional word embeddings. This goal is achieved by removing the parameter $N$ (the extent of the feature overlap to be taken into account) and adding a smoothing parameter that is proven to be constant under multiple settings, therefore making the measure unsupervised.

Our experiments show performance improvements both in *similarity estimation* and in the more challenging *outlier detection* task (Camacho-Collados and Navigli, 2016), which consists in cluster and outlier identification.[2]

---

[1] Enrico Santus and Hongmin Wang equally contributed to this work, which was started while they were both affiliated to the Singapore University of Technology and Design.

[2] Code and vectors used for the experiments are available at https://github.com/esantus/Outlier_Detection.

## 2 Similarity, Relatedness and Dissimilarity: Current Issues in the Evaluation of DSMs

A classical benchmark for DSMs is represented by the estimation of word similarity: evaluation datasets are built by asking human subjects to rate the degree of semantic similarity of word pairs, and the performance is assessed in terms of the correlation between the average scores assigned to the pairs by the subjects and the cosines of the corresponding vectors (*similary estimation* task).

Similarity as modeled by DSMs has been under debate, as its definition is underspecified. It in fact includes an ambiguity with the more generic notion of *semantic relatedness*, which is present also in many popular datasets (i.e. the concepts of *coffee* and *cup* are certainly related, but there is very little similarity about them), as opposed to 'genuine' *semantic similarity* (i.e. the relation holding between concepts such as *coffee* and *tea*) (Agirre et al., 2009; Hill et al., 2015; Gladkova and Drozd, 2016). Therefore, when testing a DSM, it is important to pay attention to what type of semantic relation is actually modeled by the evaluation dataset. Moreover, researchers pointed out that similarity estimation alone does not constitute a strong benchmark, as the inter-annotator agreement is relatively low in all datasets and the performance of several automated systems is already above the upper bound (Batchkarov et al., 2016). As a consequence, workshops such as RepEval have been organized with the explicit purpose of finding alternative evaluation tasks for DSMs.

A recent proposal is the challenging *outlier detection* task (Camacho-Collados and Navigli, 2016; Blair et al., 2016), which consists in the recognition of *cluster membership*, as well as of a relative degree of *semantic dissimilarity*. The task is described as follows: given a group of words, identify the outlier, namely the word that does not belong to the group (i.e. the one that is less similar to the others). On top of its potential applications (e.g. ontology learning), detecting outliers in clusters is a goal that poses a more strict quality requirement on the distributional representations compared to tests based simply on pairwise comparisons, as it is required that similar words group into semantically meaningful clusters. Clearly, the task involves the identification of discriminative semantic dimensions, which could set the cluster members apart from non-members. Outliers are not necessarily unrelated to the other words: rather, they have a lower degree of similarity with respect to some prominent property of the cluster (e.g. the case of *Los Angeles Lakers* as an outlier in a cluster of *football teams*). In our view, a similarity metric has to exploit such discriminative dimensions to form cohesive clusters.

## 3 A Rank-Based Metric for Embeddings

We use cosine as a baseline and we test an adaptation of a rank-based measure to the dense features of the word embeddings.

Vector cosine computes the correlation between all the vector dimensions, independently of their relevance for a given word pair or for a semantic cluster, and this could be a limitation for discerning different degrees of dissimilarity. The alternative rank-based measure is based on the hypothesis that similarity consists of sharing many relevant features, whereas dissimilarity can be described as either the non-sharing of relevant features or the sharing of non-relevant features (Santus et al., 2014, 2016b).

This hypothesis could turn out to be very helpful for a task like the outlier detection, where prominent features might be the key to improve clustering quality: semantic dimensions that are shared by many of the cluster elements should be weighted more, as they are likely to be useful for setting the outliers apart. In fact, a cohesive cluster should be mostly characterized by the same 'salient' dimensions, and thus, basing word comparisons on such dimensions should lead to more reliable estimates of cluster membership.

In our contribution, we propose to adapt $APSyn$, a metric originally proposed by Santus et al. (2016a,b), to dense word embeddings representations.[3] $APSyn$ was shown to perform well on both synonymy detection and similarity estimation tasks, and it was recently adapted to achieve state-of-the-art results in thematic fit estimation (Santus et al., 2017). The original $APSyn$ formula is shown in equation 1.

$$APSyn(w_x, w_y) = \sum_{i=0}^{i=N} \frac{1}{AVG(r_{s_x}(f_i), r_{s_y}(f_i))} \quad (1)$$

For every feature $f_i$ in the intersection between the top $N$ features of two vectors $w_x$ and $w_y$, we add the inverse of the average rank of such feature, $r_{s_x}(f_x)$ and $r_{s_y}(f_y)$, in the two decreasingly value-sorted vectors $s_x$ and $s_y$ (in traditional vectors, often the parameter $N \geq 1000$, but in WE $N = |f|$). $APSyn$ scores low if the features of the two vectors are inversely ranked and high if they are similarly ranked.

---
[3]The number of dimensions in word embeddings is in the scale of hundreds, and thus the dimensionality is way lower than in the original DSMs used by Santus and colleagues.

$APSyn$ maps the average feature ranks to a non-linear function, emphasizing the contribution of top-ranked features. Its direct application to dense embeddings would shrink too much the contribution of lower ranks (see Figure 1), with the score mostly affected by the top $\sim 25$ features. While this is reasonable for the traditional vectors derived from co-occurrence counts, where thousands of smaller contributions can still affect the final score, dense vectors need a smoother curve. While preserving the idea of the non-linear weight allocation across the average feature ranks during the summation, we modify the original $APSyn$ formula by taking the exponential of the feature ranks to a power of a constant value ranging between 0 and 1 (excluded), as shown in equation 2, such that now the number of ranks contributing to the final score is widen to all features (see the smoother curve of $APSynPower$ in Figure 1). We name this variant $APSynPower$ or, shortly, $APSynP$.

$$APSynP(w_x, w_y) = \sum_{i=0}^{i=|f|} \frac{1}{AVG(r_{s_x}(f_i)^p, r_{s_y}(f_i)^p))}$$
(2)

The power $p$ added to $APSynP$ formula is a trainable parameter. We trained it on the similarity subset of WordSim dataset, obtaining the optimal value of $p = 0.1$, which has been successfully used in all evaluations, under all settings (i.e. embedding types and training corpora). Such regularity allows us to consider $p = 0.1$ as a constant, therefore dropping $p$. Since in WE we can drop also the parameter $N$ by defining $N = |f|$, $APSynP$ can be not parametrized at all.

## 4 Evaluation Settings

### 4.1 Embeddings

For our experiments, we used two popular word embeddings architectures: the Skip-Gram with negative sampling (Mikolov et al., 2013a,b) and the GloVe vectors (Pennington et al., 2014) (standard hyperparameter settings: 300 dimensions,

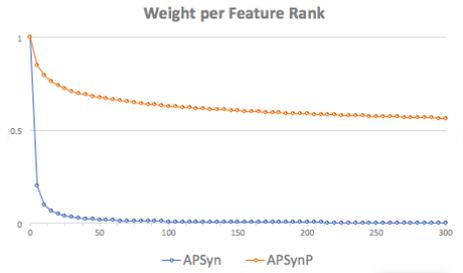

Figure 1: Comparison of weight per feature rank in $APSyn$ and $APSynP$ ($p = 0.1$) across feature ranks ranging from 1 to 300.

context size of 10 and negative sampling).[4]

For comparison with Camacho-Collados and Navigli (2016) on outlier detection, we used the same training corpora: the UMBC (Han et al., 2013) and the English Wikipedia.[5]

### 4.2 Datasets

As for the similarity estimation task, we evaluate the Spearman correlation between system-generated scores and human judgments. We used three popular benchmark datasets: WordSim-353 (Finkelstein et al., 2001), MEN (Bruni et al., 2014) and SimLex-999 (Hill et al., 2015). It is important to point out that SimLex-999 is the only one specifically built for targeting genuine semantic similarity, while the others tend to mix similarity and relatedness scores.

As for outlier detection, we evaluate our DSMs on the 8-8-8 dataset (Camacho-Collados and Navigli, 2016). The dataset consists of eight clusters, each one with a different topic and consisting in turn of eight lexical items belonging to the cluster and eight outliers (with four degrees of relatedness to the cluster members: C1, C2, C3, C4). In total, the dataset includes 64 sets of 8 words + 1 outlier for the evaluation. For each word $w$ of a cluster $W$ of $n$ words, the authors defined a compactness score $c(w)$ corresponding to the average of all pairwise similarities of the words in $W \setminus \{w\}$. On the basis of the compactness score, they proposed two evaluation metrics: Outlier Position (OP) and Outlier Detection (OD). Given a set $W$ of $n + 1$ words, OP is the rank of the outlier $w_n + 1$ according to the compactness score. Ideally, the rank of the outlier should be $n$, mean-ing that it has the lowest average similarity with the other cluster elements. The second metric, Outlier Detection (OD), is indeed defined as 1 iff $OP(w_n + 1) = n$, 0 otherwise. Finally, the performance on a dataset composed of $|D|$ sets of words was estimated in terms of *Outlier Position Percentage* ($OPP$, Eq. 3) and $Accuracy$ (Eq. 4):

$$OPP = \frac{\sum_{W \in D} \frac{OP(W)}{|W|-1}}{D} \times 100 \qquad (3)$$

$$Accuracy = \frac{\sum_{W \in D} OD(W)}{D} \times 100 \qquad (4)$$

---
[4] We also performed experiments with CBOW embeddings (Mikolov et al., 2013b), but results were irregular and inconsistent. We leave therefore their analysis to future work.
[5] Dump of Nov. 2014, approx. 1.7 billion words.

|  | Skip-Gram | | | GloVe | | |
|---|---|---|---|---|---|---|
|  | WordSim-353 | MEN | Simlex-999 | WordSim-353 | MEN | Simlex-999 |
| $Cosine$ | **0.736** | **0.758** | 0.364 | 0.511 | 0.640 | 0.311 |
| $APSyn$ | 0.599 | 0.643 | 0.343 | 0.356 | 0.393 | 0.174 |
| $APSynP$ | 0.710 | 0.737 | **0.369** | **0.607** | **0.670** | **0.335** |

Table 1: Similarity Estimation, Spearman Correlation by Setting. Embeddings trained on Wikipedia.

|  | Skip-Gram | | | | GloVe | | | |
|---|---|---|---|---|---|---|---|---|
|  | UMBC | | Wiki | | UMBC | | Wiki | |
|  | OPP | Acc. | OPP | Acc. | OPP | Acc. | OPP | Acc. |
| $CC - Cos$ | 92.6 | 64.1 | 93.8 | 70.3 | 81.6 | 40.6 | 91.8 | 56.3 |
| **Pairwise** | | | | | | | | |
| $APSyn$ | 93.0 | 67.2 | 94.0 | 68.8 | 78.7 | 40.6 | 89.3 | 53.1 |
| $APSynP$ | **94.0** | **68.8** | **94.5** | **73.4** | **81.8** | **42.2** | **92.8** | **61.0** |
| **Prototype** | | | | | | | | |
| $PT - Cos$ | 93.4 | 65.6 | 93.8 | 68.8 | 80.3 | 40.6 | 90.6 | 54.7 |
| $APSyn$ | 92.6 | 70.3 | 91.0 | 62.5 | 81.6 | 40.6 | 88.7 | 54.7 |
| $APSynP$ | **94.0** | **70.3** | **94.9** | **73.4** | **82.2** | **43.8** | 92.0 | 60.9 |

Table 2: Outlier Detection, Performance by Setting. *CC-Cos* refers to Camacho-Collados and Navigli (2016)'s pairwise method, while *PT-Cos* refers to the prototype-based one. In bold, best scores per method; in bold and underlined, best scores per corpus-embedding combination.

### 4.3 Pairwise and Prototype Approaches to Outlier Detection

While for the similarity task scores are always calculated pairwise, for spotting the outlier two different methods were tested: the *pairwise comparisons* and the *cluster prototype*.

In the first case, we reimplemented the method of Camacho-Collados and Navigli (2016): (i) compute the average similarity score of each word with the other words in the cluster; (ii) pick as the outlier the word with the lowest average score. An alternative consists in creating a cluster prototype: (i) for a cluster of N words, we create N prototype vectors by excluding each time one of the words and averaging the vectors of the other ones; (ii) we pick as the outlier the word with the lowest similarity score with the prototype built out of the vectors of the other words in the cluster.

## 5 Results and Discussion

Table 1 summarizes the correlations for the similarity task. $APSynP$ outperforms both vector cosine and $APSyn$ in all the datasets described in 4.2 when GloVe embeddings are used. The advantage is statistically significant over the cosine on the MEN dataset ($p < 0.05$) and over $APSyn$ on all datasets ($p < 0.01$).[6] With Skip-Gram embeddings, $APSynP$ performs comparably to vector cosine for relatedness, dominant in WordSim and MEN, while retaining a significant advantage over $APSyn$ on the same datasets ($p < 0.05$). It also performs slightly better than cosine in SimLex-999, and this complies with previous findings of Santus et al. (2016a), who showed that $APSyn$ performs better on genuine similarity datasets. Apparently, the top-ranked vector dimensions (those contributing more to APSyn scores) are more often shared by similar words, than by simply related ones.

Table 2 shows the results for the outlier detection task. The line *CC-Cos* contains the scores by Camacho-Collados and Navigli (2016) as a baseline. The models are divided into *pairwise comparison* and *cluster prototype* (see Section 4.3).

As it can be easily noticed by looking at the bold line, $APSynP$ outperforms the baselines in all settings for both Skip-Gram and GloVe, obtaining higher accuracies and OPPs. Not only $APSynP$ is better at identifying the outlier, but when it is not able to do so, its error is minimum (e.g. the outlier is eventually the second ranked candidate). The best accuracy (73.4 vs. SOA of 70.3) and the best OPP (94.9 vs. SOA of 93.8) are both obtained by $APSynP$ with the prototype approach, using the Skip-Gram trained on Wikipedia. We also tested the significance of the accuracy improvements with the $\chi^2$ test but, also given the small size of the 8-8-8 dataset, the result was negative.

---
[6]p-values computed with Fisher's r-to-z transformation.

Finally, we observe that the two approaches described in 4.3 do not lead to sensitively different results. The major factors of difference can be found instead in the embeddings (with Skip-Gram outperforming Glove) and in the training corpus (the smaller Wikipedia, 1.7B words, outperforms the bigger UMBC, 3B words).

### 5.1 Error Analysis

In Table 3, we report the 5 outliers that were most difficult to detect by $APSynP$. Most of them are related to the *German Car Manufacturers* topic, which was ambiguous and populated by rare terms. All outliers in the *Months* and in the *South American countries* clusters (except for the two South-American cities *Rio de Janeiro* and *Bogotá*) are successfully identified under all experimental settings. Finally, the reader can notice that most errors belong to C1 and C2, which are the most challenging classes in the dataset, as these outliers are either very related or very similar to other cluster members.

| Cluster | Outlier | Class |
|---|---|---|
| GCM | Bridgestone | C1 |
| AJC | Mary | C1 |
| GCM | Michael Schumacher | C3 |
| SSP | Moon | C1 |
| GCM | Samsung | C2 |
| BC | dolphin | C2 |
| ITC | software | C3 |
| BC | dog | C1 |
| GCM | Michelin | C1 |
| ITC | Adidas | C2 |

Table 3: Outlier Detection: Top 10 common errors across settings and their difficulty class (i.e. C1, C2, C3 and C4). (GCM: German Car Manufacturers; AJC: Apostles of Jesus Christ; SSP: Solar System Planets; BC: Big Cats; ITC: IT Companies).

## 6 Conclusions

We have introduced $APSynP$, an adaptation of the rank-based similarity measure $APSyn$ (Santus et al., 2016a,b) for word embeddings. This adaptation introduces a power parameter $p$, which is shown to be constant in multiple tasks (i.e. $p = 0.1$). The stability of this parameter, together with the possibility of dropping the parameter $N$ of $APSyn$ when using WE by setting $N = |f|$, makes the measure unsupervised. We have tested it on the tasks of *similarity estimation* and *outlier detection*, obtaining similar or better performances than vector cosine and the original $APSyn$. $APSynP$ performs more consistently on SimLex-999, showing a preference for genuine similarity, as already noticed by Santus et al. (2016a). We also introduced a new approach to the outlier detection task, based on a cluster prototype. The prototype method is competitive and computationally less expensive than pairwise comparisons.

We leave to future work a systematic comparison of $APSynP$ and other rank-based measures. Pilot tests have shown that other rank-based metrics (e.g. Spearman's Rho) also outperform vector cosine in multiple settings and tasks.

## 7 Acknowledgments

The authors thank the reviewers for the constructive reviews.

Enrico Santus' research is supported by the Singapore University of Technology and Design (SUTD) and by the Massachusetts Institute of Technology (MIT).

Emmanuele Chersoni's research is supported by an A*MIDEX grant (nANR-11-IDEX-0001-02), funded by the French Government "Investissements d'Avenir" program.